\documentclass[letterpaper, 10 pt, conference]{IEEEtran}  % Comment this line out if you need a4paper

\IEEEoverridecommandlockouts                              % This command is only needed if 
\usepackage{graphicx} % for pdf, bitmapped graphics files
\usepackage{caption}
\usepackage{enumerate}
\usepackage{subcaption}
\usepackage{booktabs}       % professional-quality tables
\usepackage{siunitx}
\usepackage{bm} % Package for bold symbols in math mode
\usepackage{hyperref}
\usepackage{amsmath}
\usepackage[capitalise]{cleveref}
\usepackage[acronym,shortcuts]{glossaries}
\usepackage[numbers,sort]{natbib}
\usepackage{algorithm}
\usepackage{algpseudocode}
\usepackage[table,xcdraw]{xcolor}
\usepackage{cuted}
\usepackage{amssymb} % For the \checkmark symbol
\usepackage{pifont}  % For the \xmark symbol
\usepackage{adjustbox} % for adjust box
\usepackage{stfloats} % for bottom table* to work

\newacronym{ads}{ADS}{Autonomous Driving Systems}
\newacronym{nn}{NN}{Neural Network}
\newacronym{ros}{ROS}{Robot Operating System}
\newacronym{sota}{SotA}{State-of-the-Art}
\newacronym{rcarla}{$\mathcal{R}$-CARLA}{$\mathcal{R}$acing CARLA}
\newacronym{npc}{NPC}{Non-Player Character}
\newacronym{imu}{IMU}{Inertial Measurement Unit}
\newacronym{lidar}{LiDAR}{Light Detection and Ranging}
\newacronym{rmse}{RMSE}{Root Mean Squared Error}
\newacronym{slam}{SLAM}{Simultaneous Localization and Mapping}

\newcommand{\xmark}{\ding{55}} % defines x mark, uses pifont

\newcommand{\rcarla}{\gls{rcarla}}

\usepackage{eso-pic}

\newcommand\AtPageUpperCenterNotice[1]{%
  \AtPageUpperLeft{%
    \put(\LenToUnit{0.5\paperwidth},\LenToUnit{-2cm}){\makebox[0pt]{#1}}%
  }%
}

\AddToShipoutPictureBG*{%
  \AtPageUpperCenterNotice{%
    \parbox[b][2cm][c]{\paperwidth}{%
      \centering
      \fontsize{12}{14}\selectfont
      \color{gray!50}
      This paper has been accepted for publication at the\\
      Intelligent Vehicles Symposium (IV), Romania 2025. \copyright{}IEEE
    }%
  }%
}

\AddToShipoutPictureBG*{
  \AtPageLowerLeft{%
    \raisebox{25pt}{\makebox[\paperwidth]{\begin{minipage}{21cm}\centering
    \fontsize{10}{12}\selectfont
    \textcolor{gray!50}{%
      \copyright{} 2025 IEEE. Personal use of this material is permitted.
      Permission from IEEE must be obtained for all other uses, in any current or future media,
      including reprinting/republishing this material for advertising or promotional purposes,
      creating new collective works, for resale or redistribution to servers or lists,
      or reuse of any copyrighted component of this work in other works.
    }
    \end{minipage}}}%
  }
}

\title{\LARGE \bf
\acrshort{rcarla}: High-Fidelity Sensor Simulations with Interchangeable Dynamics for Autonomous Racing
}

\author{Maurice Brunner\IEEEauthorrefmark{1}, Edoardo Ghignone\IEEEauthorrefmark{1}, Nicolas Baumann\IEEEauthorrefmark{1}, and Michele Magno\IEEEauthorrefmark{1}% <-this % stops a space
\thanks{\IEEEauthorrefmark{1}Maurice Brunner, Edoardo Ghignone, Nicolas Baumann, and Michele Magno are associated with the Center for Project-Based Learning, D-ITET, ETH Zurich.}%
\thanks{Corresponding Author: Maurice Brunner.}%
\thanks{\tt\small maurice.brunner@pbl.ee.ethz.ch}
}

\begin{document}

\maketitle
\begin{strip}
    \vspace{-2.75cm}
    \centering
    \includegraphics[angle=0,origin=c,width=\textwidth]{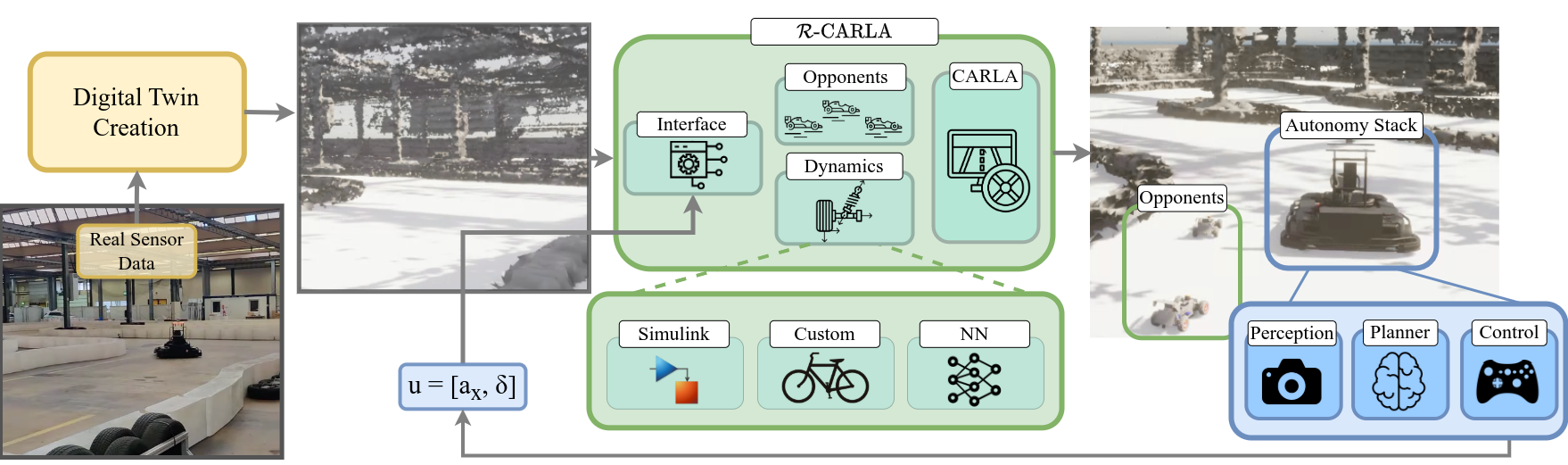}
    \captionof{figure}{Overview of the 5 main modules of \gls{rcarla}. On the left, the digital twin creation module is shown which creates new environments from real sensor data. The vehicle drive inputs $u$ from an autonomy stack are passed through an interface to the dynamics simulator, which computes the next state of the racecar. Together with the pose of the opponents coming from the opponent simulator, this is passed to CARLA for the rendering of the current frame and the sensor simulation. The sensor data are then passed back to the autonomy stack.}
    \label{fig:sys-overview}
\end{strip} %overview
\thispagestyle{empty}
\pagestyle{empty}

%%%%%%%%%%%%%%%%%%%%%%%%%%%%%%%%%%%%%%%%%%%%%%%%%%%%%%%%%%%%%%%%%%%%%%%%%%%%%%%%

\begin{abstract}
Autonomous racing has emerged as a crucial testbed for autonomous driving algorithms, necessitating a simulation environment for both vehicle dynamics and sensor behavior. Striking the right balance between vehicle dynamics and sensor accuracy is crucial for pushing vehicles to their performance limits. However, autonomous racing developers often face a trade-off between accurate vehicle dynamics and high-fidelity sensor simulations. This paper introduces \gls{rcarla}, an enhancement of the CARLA simulator that supports holistic full-stack testing, from perception to control, using a single system. By seamlessly integrating accurate vehicle dynamics with sensor simulations, opponents simulation as \glspl{npc}, and a pipeline for creating digital twins from real-world robotic data, \gls{rcarla} empowers researchers to push the boundaries of autonomous racing development. Furthermore, it is developed using CARLA's rich suite of sensor simulations. Our results indicate that incorporating the proposed digital-twin framework into \gls{rcarla} enables more realistic full-stack testing, demonstrating a significant reduction in the \emph{Sim-to-Real} gap of car dynamics simulation by 42\%  and by 82\% in the case of sensor simulation across various testing scenarios.
\end{abstract}

%%%%%%%%%%%%%%%%%%%%%%%%%%%%%%%%%%%%%%%%%%%%%%%%%%%%%%%%%%%%%%%%%%%%%%%%%%%%%%%%
\section{INTRODUCTION}
Motorsport racing, known for its high-speed and competitive nature, has emerged as an ideal test bed, allowing researchers to evaluate and refine autonomous driving capabilities under rigorous conditions that mimic real-world edge cases \cite{betz_weneed_ar, catalyst0, catalyst1, ar_survey}. This synthesis of high-performance demands and safe testing environments has given rise to various autonomous racing competitions, ranging from full-scale events like Indy Autonomous \cite{tum_fullsystem, raji2023erautopilot, kaist_fullsystem} and Formula Student Driverless \cite{amz_fullstack} to scaled-down versions such as F1TENTH \cite{baumann2024forzaeth, okelly2020f1tenth}, each contributing to the field of \gls{ads}. 

A critical component in this domain is the use of simulators, which provide cost-effective, safe, and efficient means for participants to develop and test their systems \cite{hu2024howsimulation}. Simulators enable the approximation of the \textit{See-Think-Act} cycle \cite{Siegwart2004} within a virtual environment, aiming to replicate real-world performance. However, achieving perfect simulation fidelity is infeasible --- known as the \emph{Sim-to-Real} gap. This typically leads to the compartmentalization of testing modalities. For instance, dynamics are often evaluated using modified versions of high-fidelity racing games like \emph{Assetto Corsa} or through \emph{Simulink} models, whereas perception systems are tested using recorded real data, to reproduce sensor inputs with high accuracy \cite{tum_fullsystem, raji2023erautopilot}. Hence, currently, there is no unified simulation environment that is capable of simulating different sensors and accurate vehicle dynamics for autonomous racing.

Given these limitations, our work bridges this gap by enhancing the \emph{CARLA} simulator, incorporating a plugin that allows for the integration of custom dynamics, ranging from \emph{Simulink} models and \glspl{nn}-learned dynamics to simple coded dynamics. This integration facilitates holistic testing by enabling the use of high-resolution perceptual data in conjunction with user-defined dynamics models. Additionally, we propose a streamlined workflow for converting real-world point cloud data into \emph{Unreal Engine} maps, fostering the development of digital-twin environments that mirror actual testing locations. This approach enhances the accuracy and effectiveness of simulation-based testing and development, providing a robust platform for the autonomous racing community. 
The evaluation has been performed on two physical and simulated autonomous racecars --- a scaled F1TENTH \cite{okelly2020f1tenth} platform and a gokart \cite{gokart} --- which are deployed in real-world environments with varying friction levels to record autonomous laps using a \gls{lidar}-based \gls{slam} and model-based controller. The digital twins of each environment are then created from the recorded data, enabling different simulation pipelines to be tested: one with standard CARLA dynamics and one incorporating the proposed method --- by integrating adaptable dynamics, digital twin environmental mapping, and holistic autonomy testing --- demonstrating its improved fidelity.

Key contributions of this work include: 
\begin{enumerate}[I] 
    \item \textbf{Holistic Testing:} Our enhancements to the \emph{CARLA} simulator simplify the holistic testing of autonomous systems by supporting high-fidelity dynamics within a detailed perceptual simulation environment. This approach not only enables the testing of a full-stack autonomous system but also shows a significant reduction of the discrepancy between the simulation and the real world of 81\% by testing holistically. 
    \item \textbf{Digital-Twin:} We introduce a method for transforming real-world sensor data into detailed simulation maps, enhancing the realism and applicability of virtual testing environments. By comparing the real-world sensor data to the resulting synthetic data, using a \gls{slam} system a close resemblance was found with differences of resulting \gls{rmse} differences as low as 1.74 centimeters.
    \item \textbf{\emph{Sim-to-Real} gap reduction:} Our results on two racing platforms indicate an average reduction of the \emph{Sim-to-Real} gap of 42\% for the car dynamics simulation and 81\% for sensor simulation.
    \item \textbf{Open-Source Availability:} The entire software stack, including the CARLA plugin and digital-twin workflow, is open-sourced, providing a valuable resource for the racing and research communities. The code is available at: \href{github.com/forzaeth/rcarla}{\url{github.com/forzaeth/rcarla}}. 
\end{enumerate}

\section{RELATED WORK}
CARLA \cite{Dosovitskiy17carla} is a simulation platform for autonomous driving of great interest in the research community, as not only can it be used for development, deployment, and validation, but also supports shared benchmarks \cite{chen2024e2edriving}, safety testing \cite{ramakrishna2022anticarla, nesti2024carlagear}, high-fidelity perception pipelines \cite{delapena2022avperdevkit}, and end-to-end systems \cite{xiao2022multimodal}.
Indeed, multiple works have extended CARLA with \gls{ros}-enabled interfaces, to utilize its modular sensor simulation capabilities \cite{delapena2022avperdevkit, kaljavesi2024carlaautoware}.

However, in the context of autonomous racing, CARLA falls short on the aspect of accurate dynamic simulation \cite{remonda2024assetto}, and different other lines of work have resorted to high-fidelity racing dynamics simulators \cite{remonda2024assetto, vasco2024super, fuchs2021superhuman, weiss2024deepracing}.
This choice, however, restricts researchers either in reproducibility (the platform used in \cite{vasco2024super, fuchs2021superhuman} is not publicly available) or extensibility (as modifying the assets in the original software used by \cite{remonda2024assetto, vasco2024super, fuchs2021superhuman, weiss2024deepracing} can be cumbersome or outright impossible due to closed-source software).
Therefore, different other strategies have been chosen. For example, in autonomous racing competitions, developers and researchers usually prefer to use only real data \cite{tum_fullsystem, raji2023erautopilot}, while for scaled racing platforms, a couple of fully open-source simulators have been proposed \cite{AutoDRIVE-Simulator-2021, okelly2020f1tenth, babu2020f1tenth}.
However, the need to model, identify, and continuously learn vehicle dynamics remains \cite{weg2024learning} and therefore different follow-ups have highlighted the need to utilize learning methods to improve the generalization and effectiveness \cite{davydov2024firstlearn, xiao2024anycar} of algorithms trained in simulation on the real world. 

This work proposes \rcarla, a simulator framework that comes out from the intersection of these different necessities. Firstly, it builds on top of the CARLA platform, retaining its high-fidelity sensor simulation capabilities, modular scenario construction, and generally open-source structure.
Furthermore, it extends the simulation model component to support any type of externally provided function, offering the possibility to integrate the latest methods in a setup that enables full-stack simulated testing, from perception input to control output.
Finally, multiple key features for autonomous racing research are available in \rcarla: \gls{npc} simulation is made available, offering the possibility of testing for detection systems and planning algorithms, and a pipeline for recreation of real-world scenarios via point-cloud data is further integrated, enabling the creation of a digital twin.
A full comparison of our proposed method with \gls{sota} works is available in \Cref{tab:rel_work}.

\begin{table}[ht]
    \centering
    \begin{adjustbox}{max width=\columnwidth}
    \begin{tabular}{l|cccc}
        \toprule
         & custom & modular & high-fidelity & multiple \\
         & environments & dynamics & sensors & \glspl{npc} \\
        \midrule 
        Videogame-Based &  &  &  &  \\
        Simulators \cite{remonda2024assetto, vasco2024super, fuchs2021superhuman, weiss2024deepracing} & \xmark & \xmark & \xmark & \checkmark \\
        \midrule
        Scaled Autonomous Racing  &  &  &  &  \\
        Simulators \cite{AutoDRIVE-Simulator-2021, okelly2020f1tenth, babu2020f1tenth} & \checkmark~(2D only) & \xmark & \xmark & \checkmark \\
        \midrule
        CARLA \cite{Dosovitskiy17carla} & \checkmark & \xmark & \checkmark & \checkmark \\
        \midrule
        \textbf{\rcarla~(Ours)} & \checkmark & \checkmark & \checkmark & \checkmark \\
        \bottomrule
    \end{tabular}
    \end{adjustbox}
    \caption{Comparison of autonomous racing simulators.}
    \label{tab:rel_work}
\end{table} %related works table

\section{METHODOLOGY}
The \gls{rcarla} framework enables the closed-loop simulation of a complete autonomy stack, allowing for integrated testing of state estimation, opponent detection, tracking, planning, and control algorithms. The system takes raw sensor data as input and generates vehicle control commands as output. An overview of the simulator architecture is provided in \cref{fig:sys-overview}, which highlights its five modules. The first module, CARLA, is a simulation platform designed for developing autonomous driving algorithms and is detailed in \cref{sec:carla}. \cref{sec:env} describes the digital twin creation module, which generates new simulation environments using sensor data collected from race tracks. A dynamics interface module, illustrated in \cref{fig:sys-overview}, facilitates the seamless integration of different vehicle dynamics models. Additionally, a car dynamics simulator incorporating tire force modeling is proposed in \cref{sec:single-track}. Finally, \cref{sec:opp} introduces an opponent simulation module that allows for the simulation of various opponents with customizable racing trajectories and velocities.

\subsection{CARLA Adaption}
\label{sec:carla}
\begin{figure}
    \centering
    \begin{subfigure}{0.46\linewidth}
    \includegraphics[width=\linewidth]{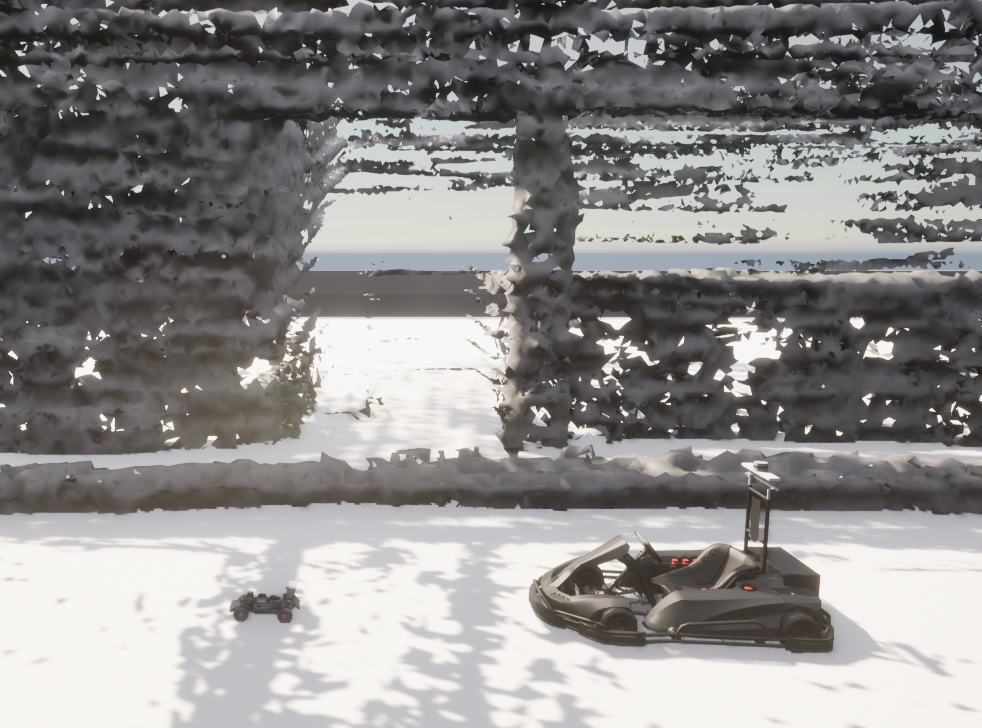}
    \caption{}
    \end{subfigure}
    \begin{subfigure}{0.46\linewidth}
    \includegraphics[width=\linewidth]{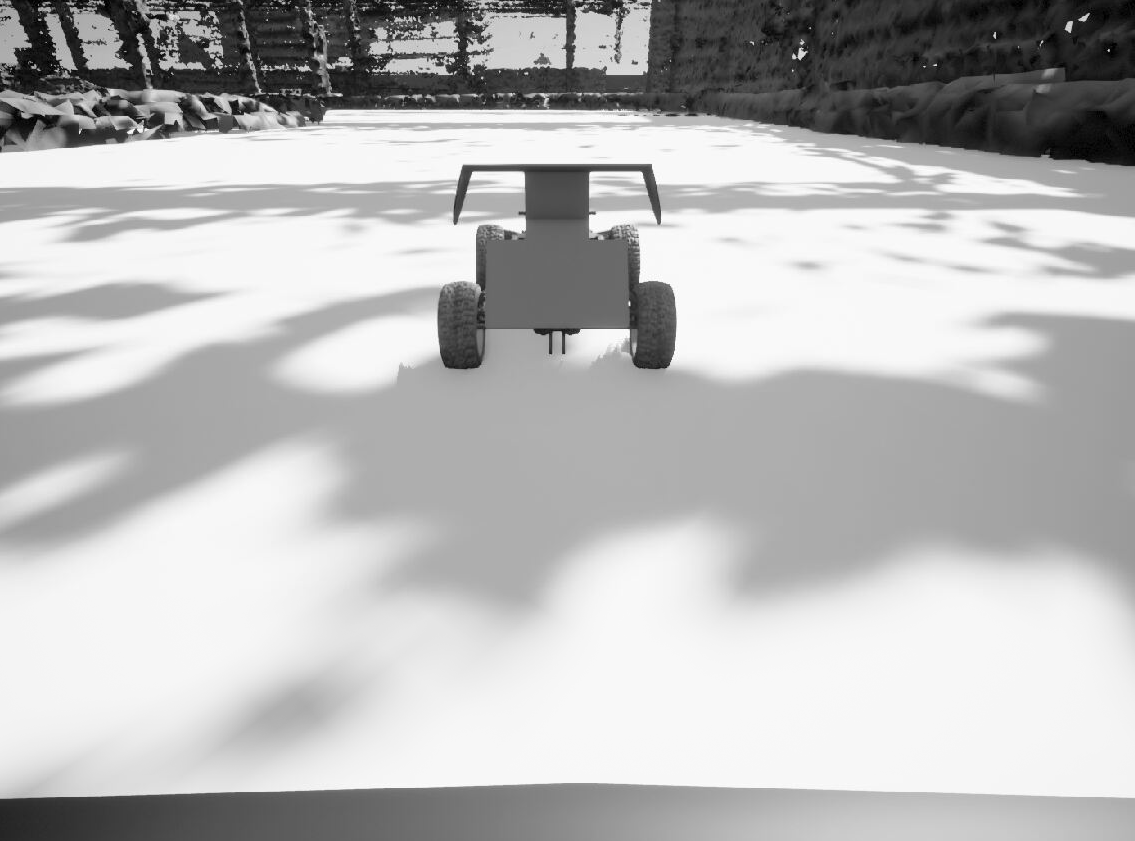}
    \caption{}
    \end{subfigure}
    \hspace{3pt}
    \begin{subfigure}{0.46\linewidth}
    \includegraphics[width=\linewidth]{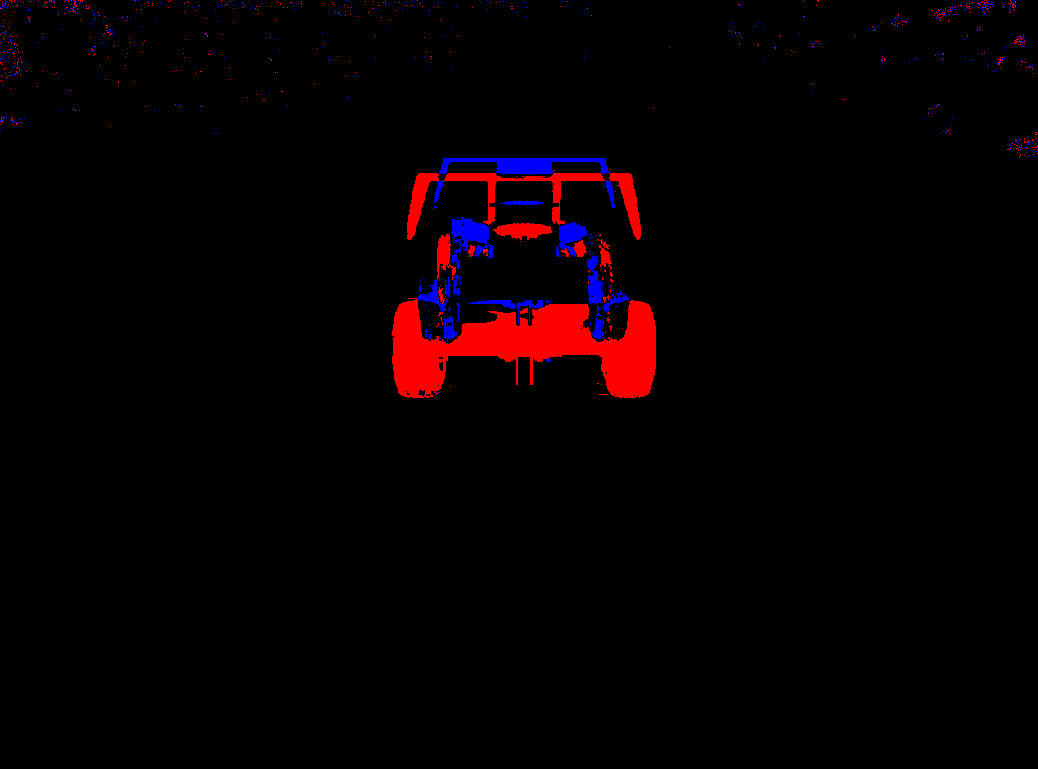}
    \caption{}
    \end{subfigure}
    \begin{subfigure}{0.46\linewidth}
    \includegraphics[width=\linewidth]{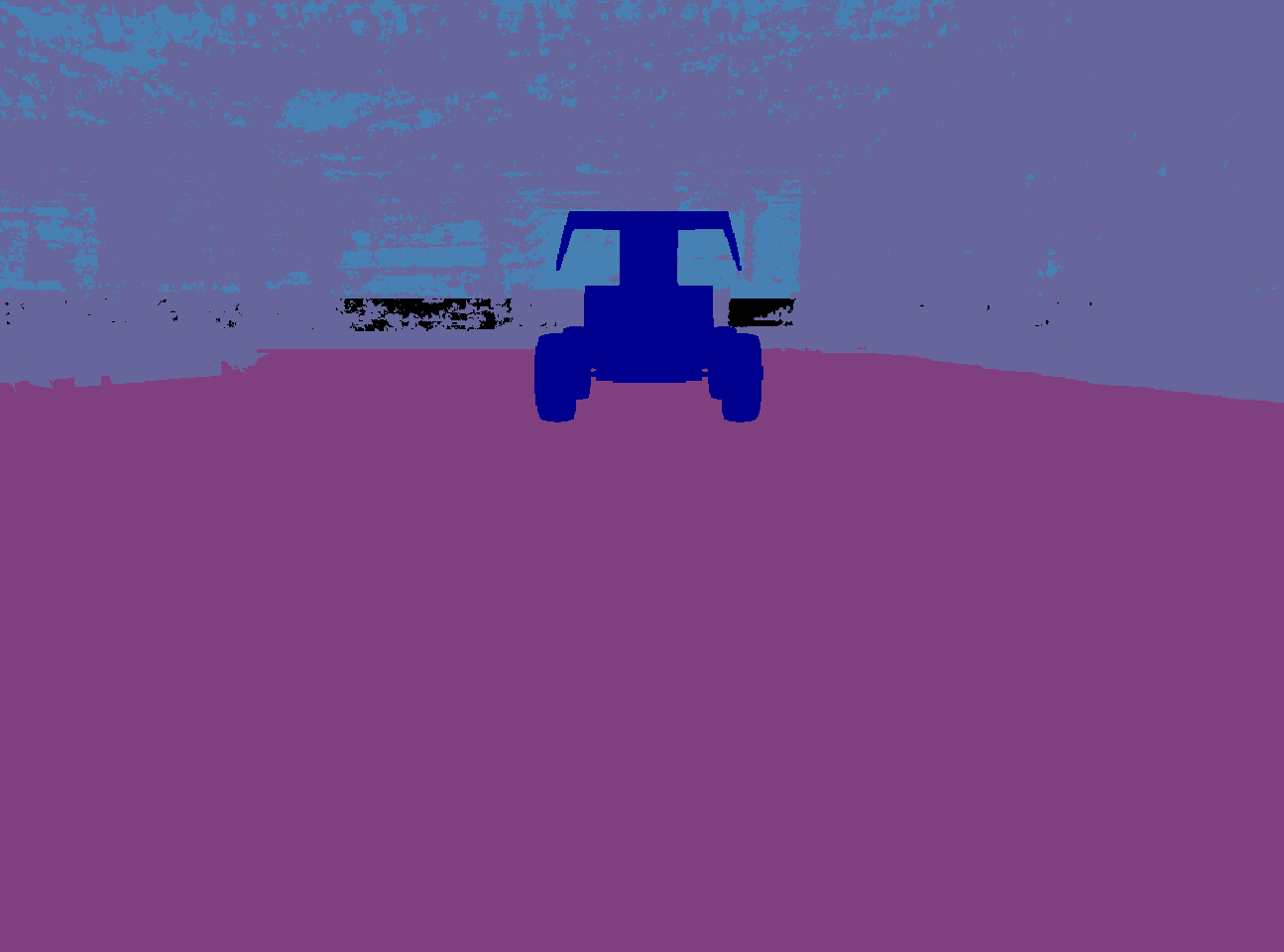}
    \caption{}
    \end{subfigure}
    \caption{Overview of different sensors and the rendering of a simulated vehicle and an opponent (a). The shown sensors are: A RGB camera (b), an event camera (b), and an image segmentation camera (c)}
    \label{fig:oppsensors}
\end{figure}

\begin{figure}
    \centering
    \begin{subfigure}{0.46\linewidth}
    \includegraphics[width=\linewidth]{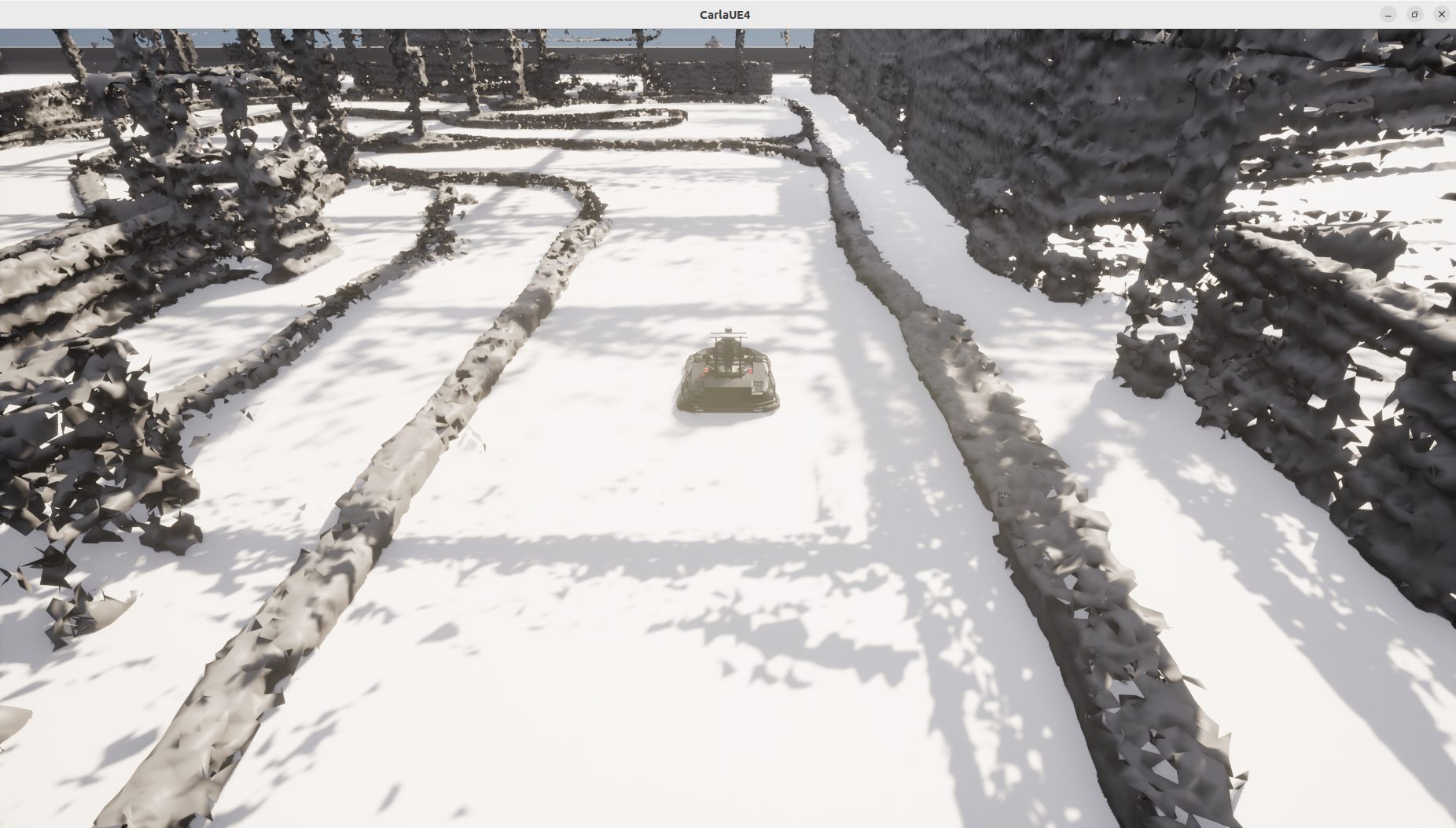}
    \caption{}
    \end{subfigure}
    \hspace{3pt}
    \begin{subfigure}{0.46\linewidth}
    \includegraphics[width=\linewidth]{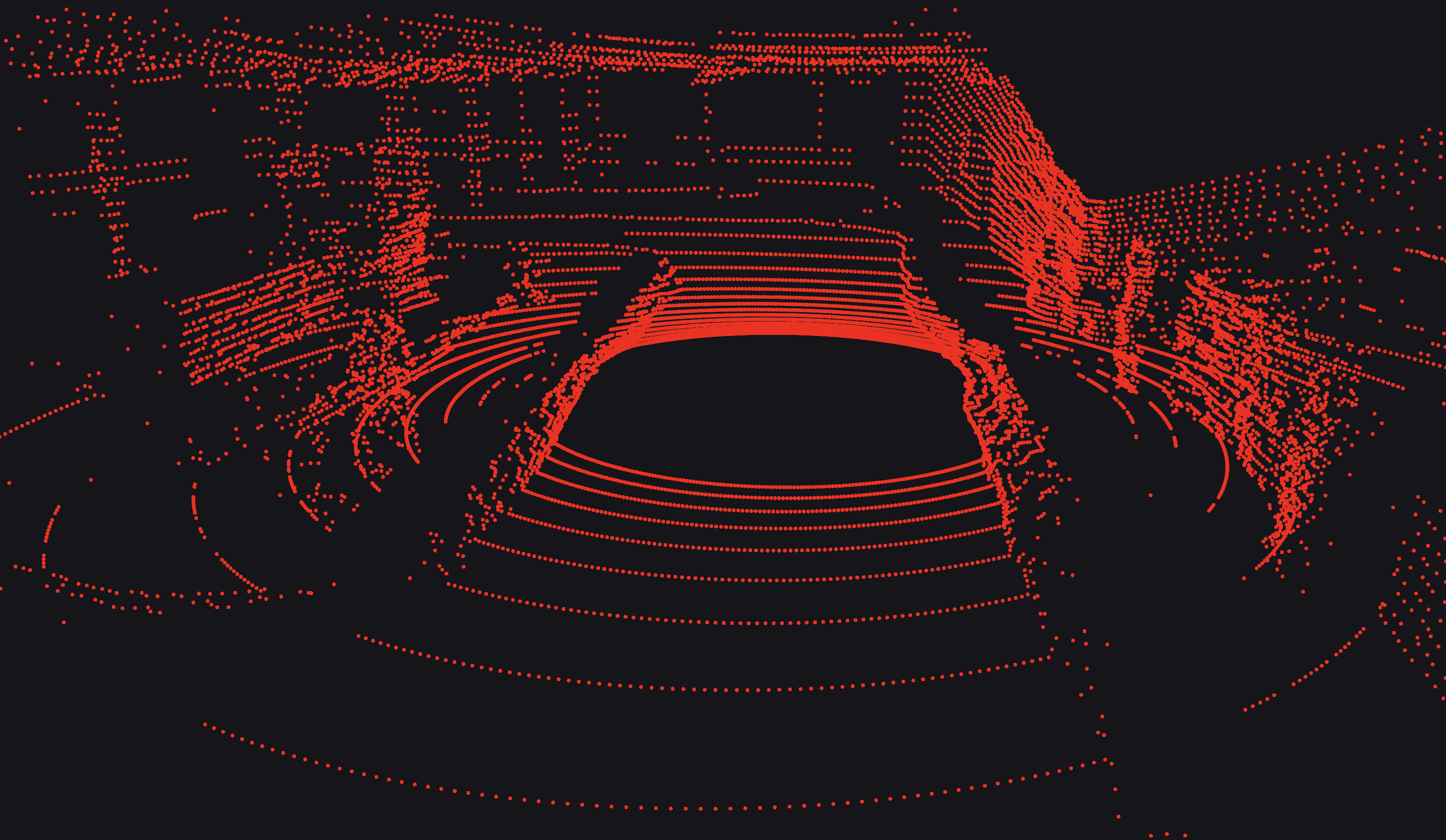}
    \caption{}
    \end{subfigure}
    % \\ \vspace{3pt}
    \begin{subfigure}{0.46\linewidth}
    \includegraphics[width=\linewidth]{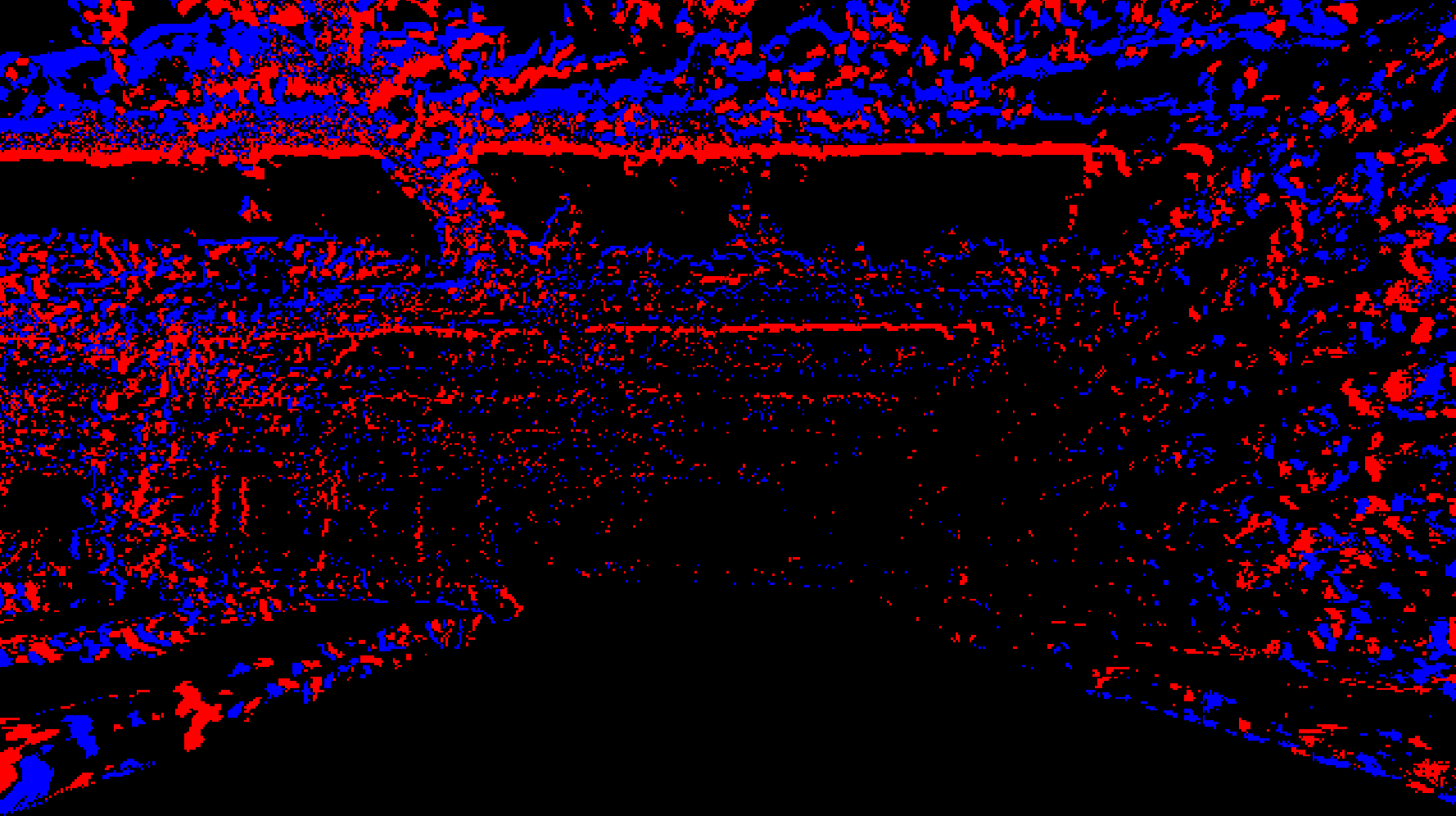}
    \caption{}
    \end{subfigure}
    \hspace{3pt}
    \begin{subfigure}{0.46\linewidth}
    \includegraphics[width=\linewidth]{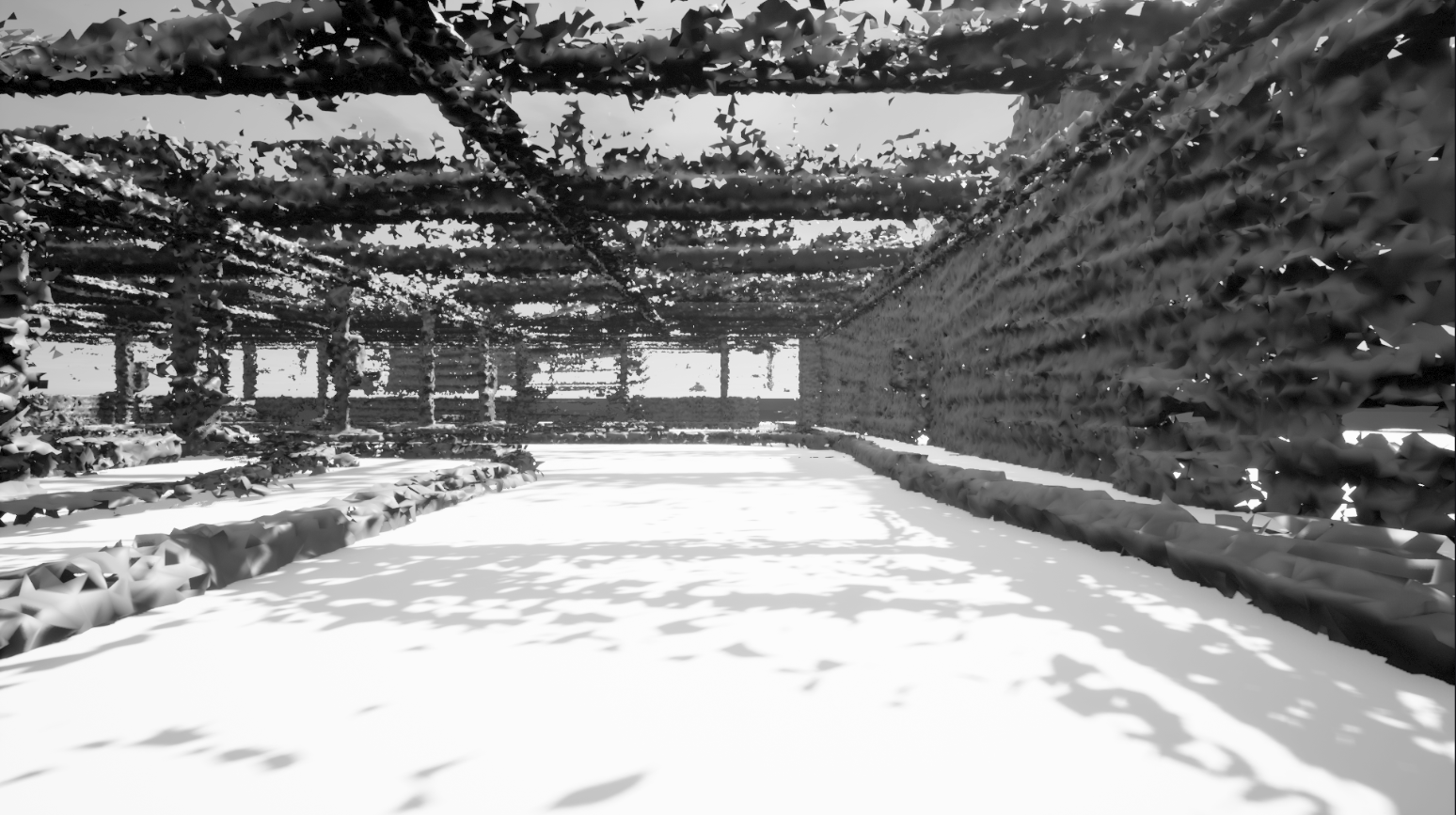}
    \caption{}
    \end{subfigure} 
    \caption{A comparison of the data of different sensors available. First (a) a picture of the simulator is shown, followed by the output of a \gls{lidar} (b), an event camera, and an image segmentation camera of the same scene.}
    \label{fig:sensors}
\end{figure}

CARLA is an open-source urban driving simulator built on the \emph{Unreal Engine}. In the context of \gls{rcarla}, the physics engine is disabled, and only the sensor simulation is utilized. At each time step $t$, the CARLA module receives updated poses for the racecar and all opponents, simulating the resulting sensor data. Additionally, it interacts with the \emph{Unreal Engine} to render the racecar and opponents. CARLA provides a wide array of sensor options, and a brief overview of these sensors is presented in \cref{fig:oppsensors} and \cref{fig:sensors}. 

\subsubsection{\gls{imu} Simulation}
In CARLA, the simulation of the \gls{imu} relies on the velocity states of the racecar. However, in \gls{rcarla}, only the pose is updated, rendering the default \gls{imu} simulation ineffective. To address this, a custom \gls{imu} simulator was developed. This new simulation computes the \gls{imu} data at time $t$ based on the state $x$ of the car at both times $t$ and $t-1$, as follows:

\begin{equation}
    \begin{bmatrix}
        a_x\\
        a_y\\
        a_z\\
        \psi\\
        \dot{\psi}
    \end{bmatrix}
    =
    \begin{bmatrix}
        \frac{v_x(t) - v_x(t-1)}{dt}\\
        \frac{v_y(t) - v_y(t-1)}{dt}\\
        g\\
        \psi(t)\\
        \dot{\psi}(t)
    \end{bmatrix}
\end{equation} %dynamics equation

 %sensors ego vehicle

 %sensors opponents

\subsection{Digital-Twin Environments}
\label{sec:env}
\begin{figure*}
    \centering
    \includegraphics[width=\textwidth]{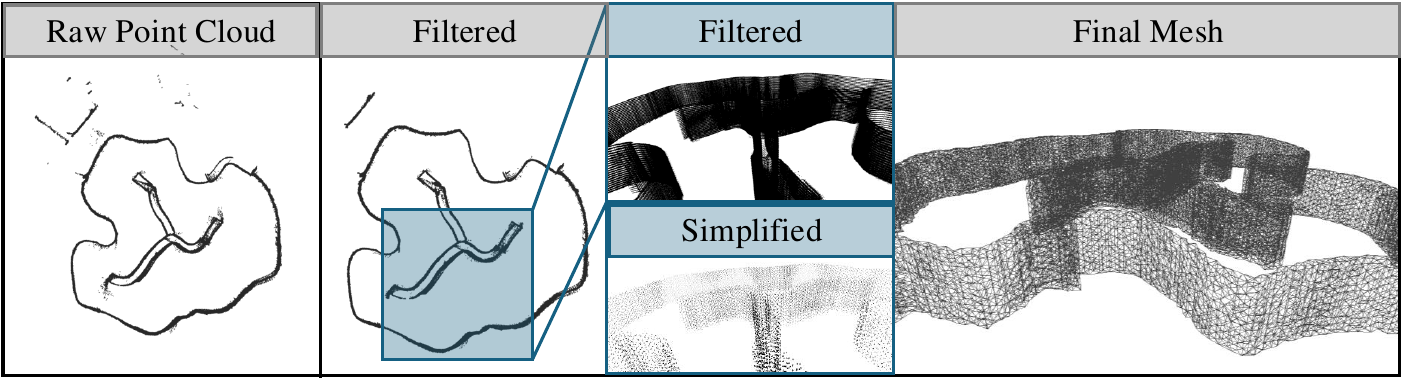}  
    \caption{Comparison of different steps in the digital-twin creation process. From left to right: The original point cloud, followed by the filtered and simplified point cloud, and the final mesh.}
    \label{fig:twincreation}
\end{figure*}

To construct a digital twin of an environment, a point cloud from real-world data is initially generated from \gls{lidar} data using Cartographer \cite{Hess2016}, however, this step can also be skipped if a point cloud is already present. The point cloud is then processed through a sphere-based outlier rejection algorithm, where points are classified as outliers if, within a sphere of radius $r$, fewer than $m$ neighboring points are present. Alternatively, a statistical outlier removal method may be employed, which filters the point cloud based on the standard deviation of the distances between neighboring points. For point clouds generated with a 2D \gls{lidar}, the data is extended along the z-axis to transform it into a 3D point cloud. The point cloud is further simplified using a Poisson-disk sampling method \cite{ulichney1988dithering}. Subsequently, a mesh is generated from the simplified point cloud using the ball-pivoting algorithm \cite{ballpivot}, which can be imported into the \emph{Unreal Engine} editor to create a new map. This process is shown in \cref{fig:twincreation}. The algorithm reduces the original pointcloud with 1750112 to 25898 points and 47864 faces.

 %twin creation

\begin{figure*}
    % \resizebox{\textwidth}{!}
    \centering
    \begin{subfigure}{0.19\linewidth}
    \includegraphics[width=\linewidth]{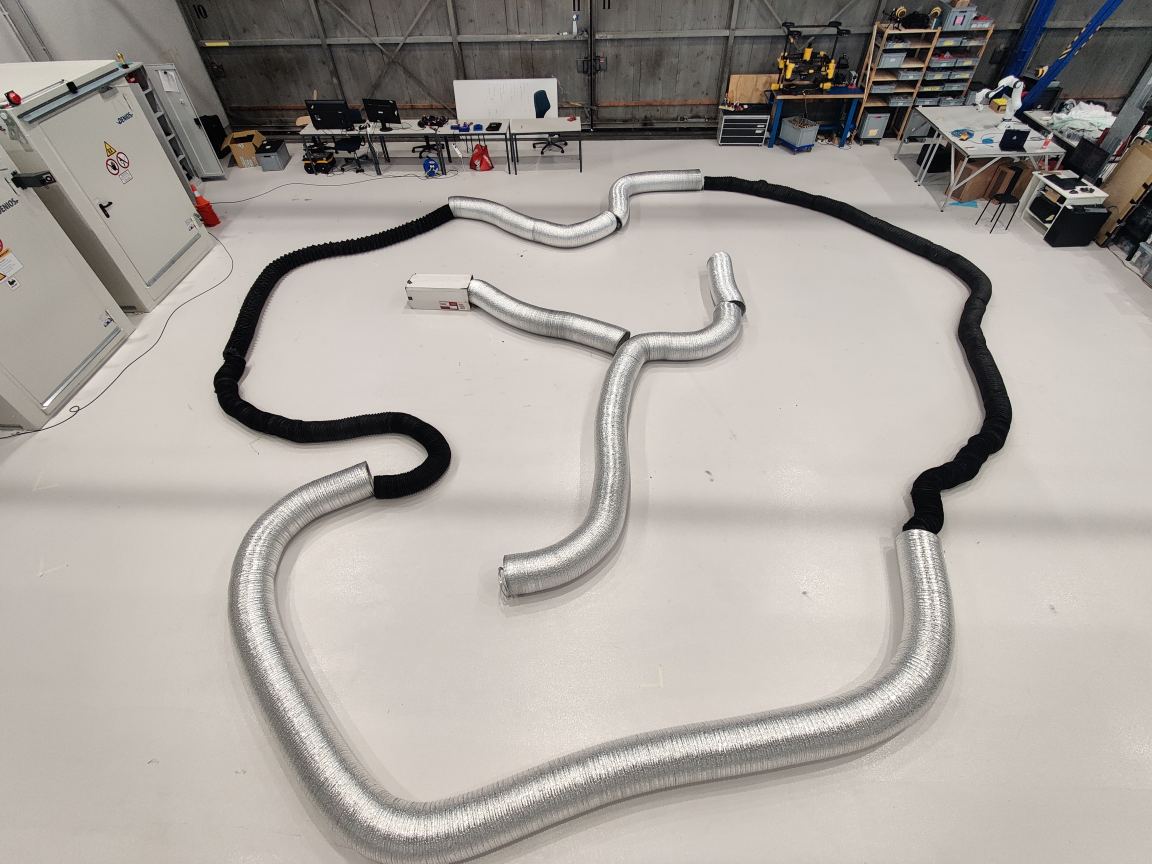}
    \end{subfigure}
    \begin{subfigure}{0.19\linewidth}
    \includegraphics[width=\linewidth]{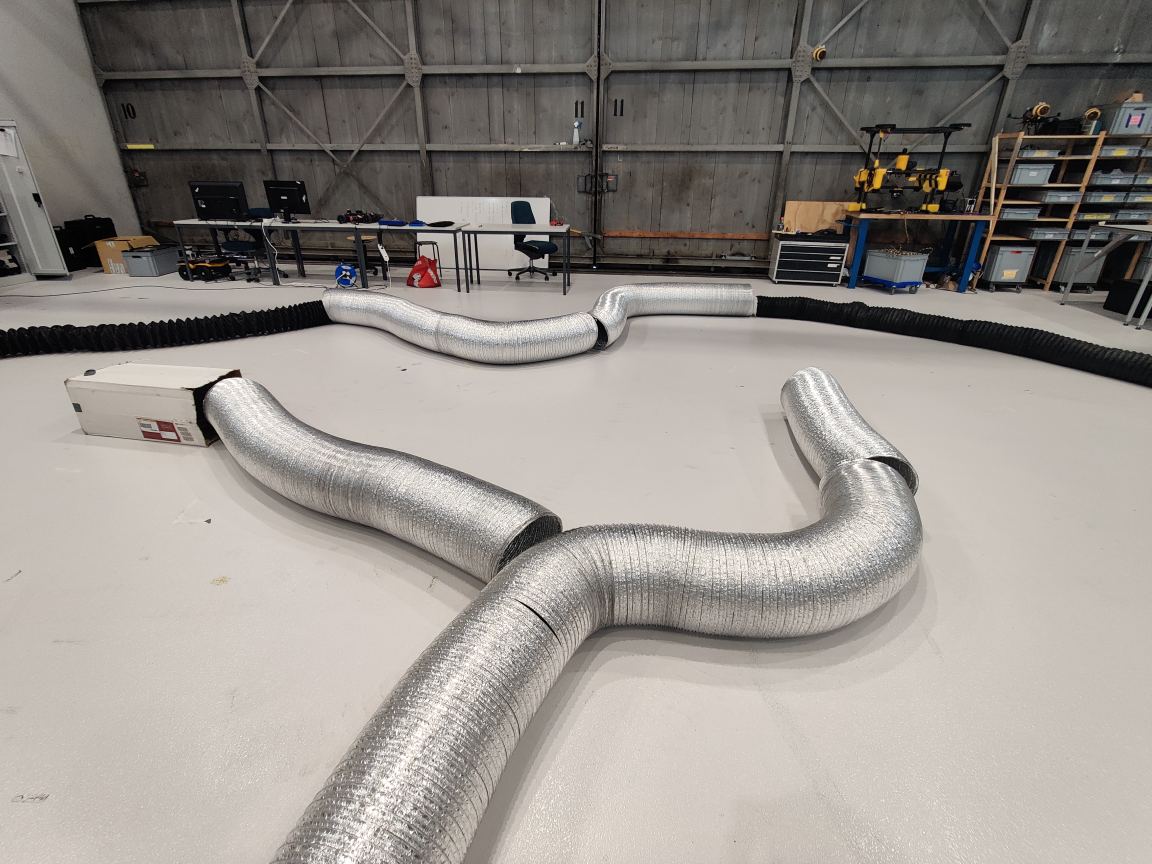}
    \end{subfigure}
    \begin{subfigure}{0.19\linewidth}
    \includegraphics[width=\linewidth]{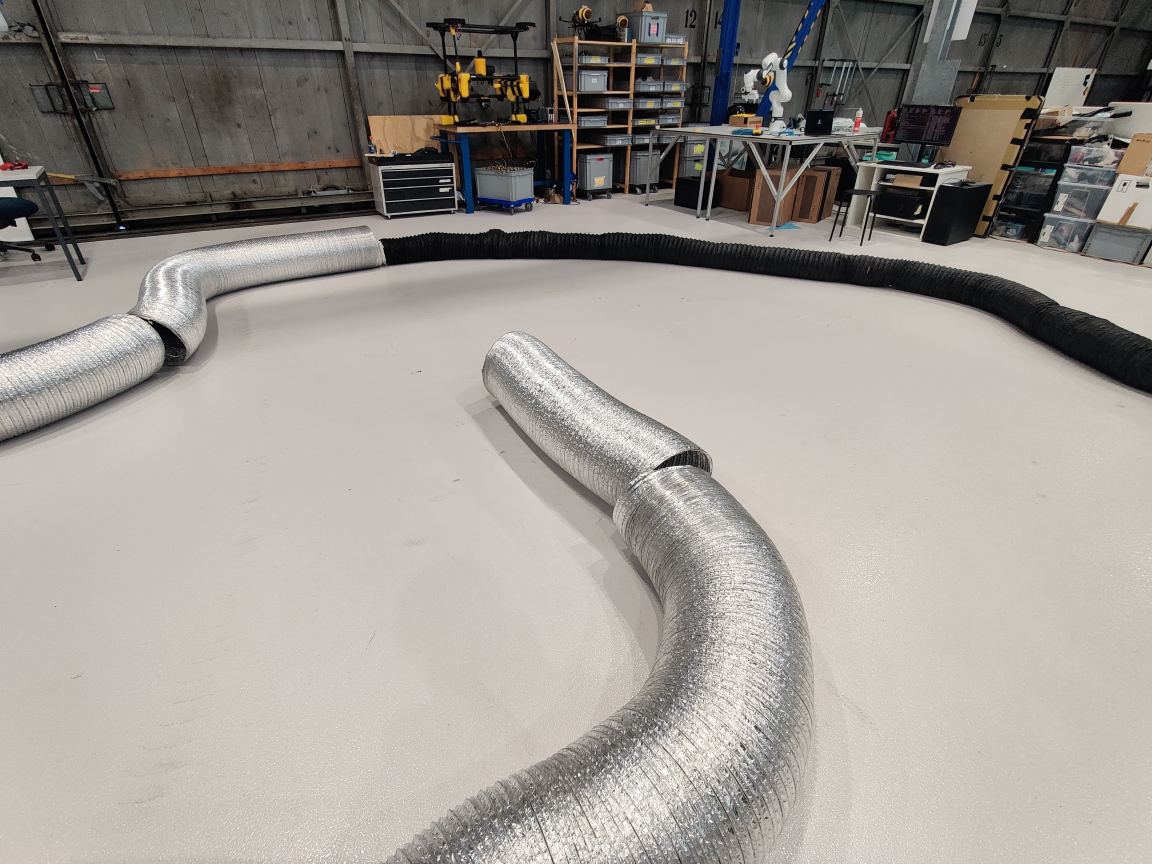}
    \end{subfigure}
    \begin{subfigure}{0.19\linewidth}
    \includegraphics[width=\linewidth]{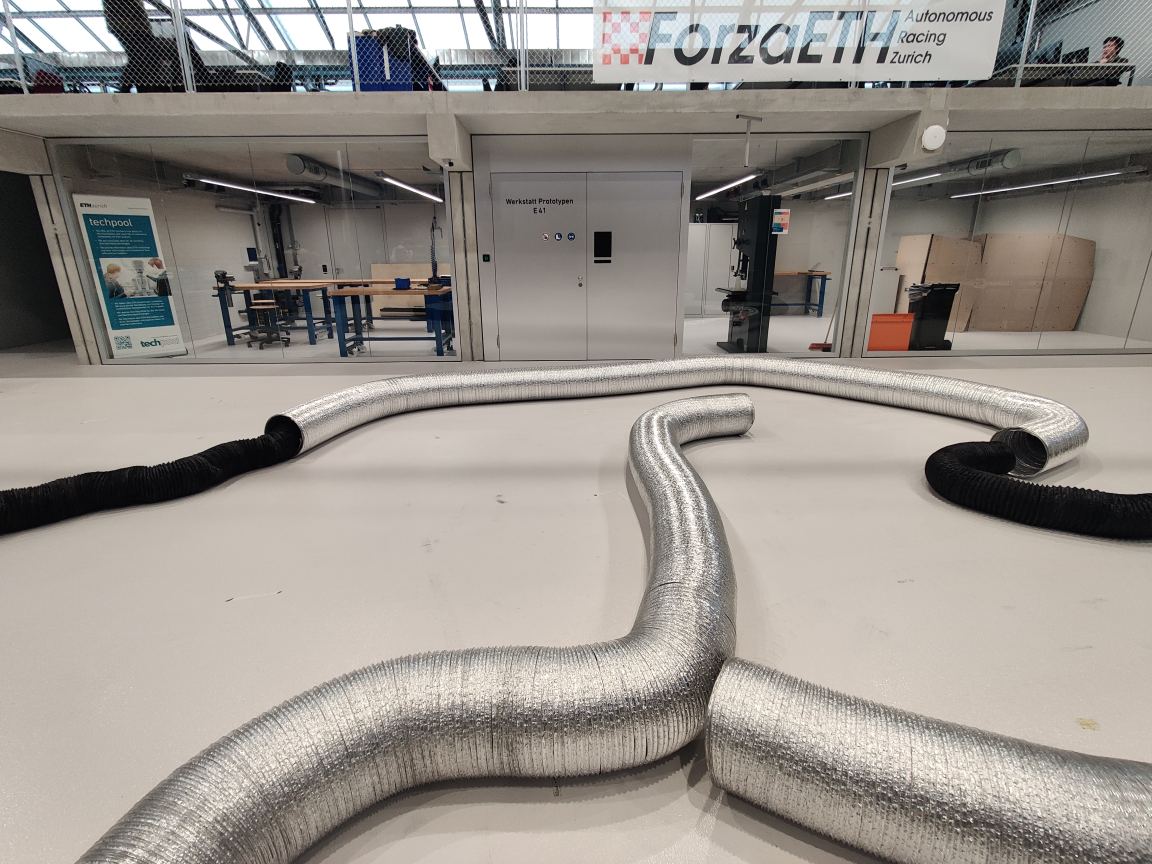}
    \end{subfigure}
    \begin{subfigure}{0.19\linewidth}
    \includegraphics[width=\linewidth]{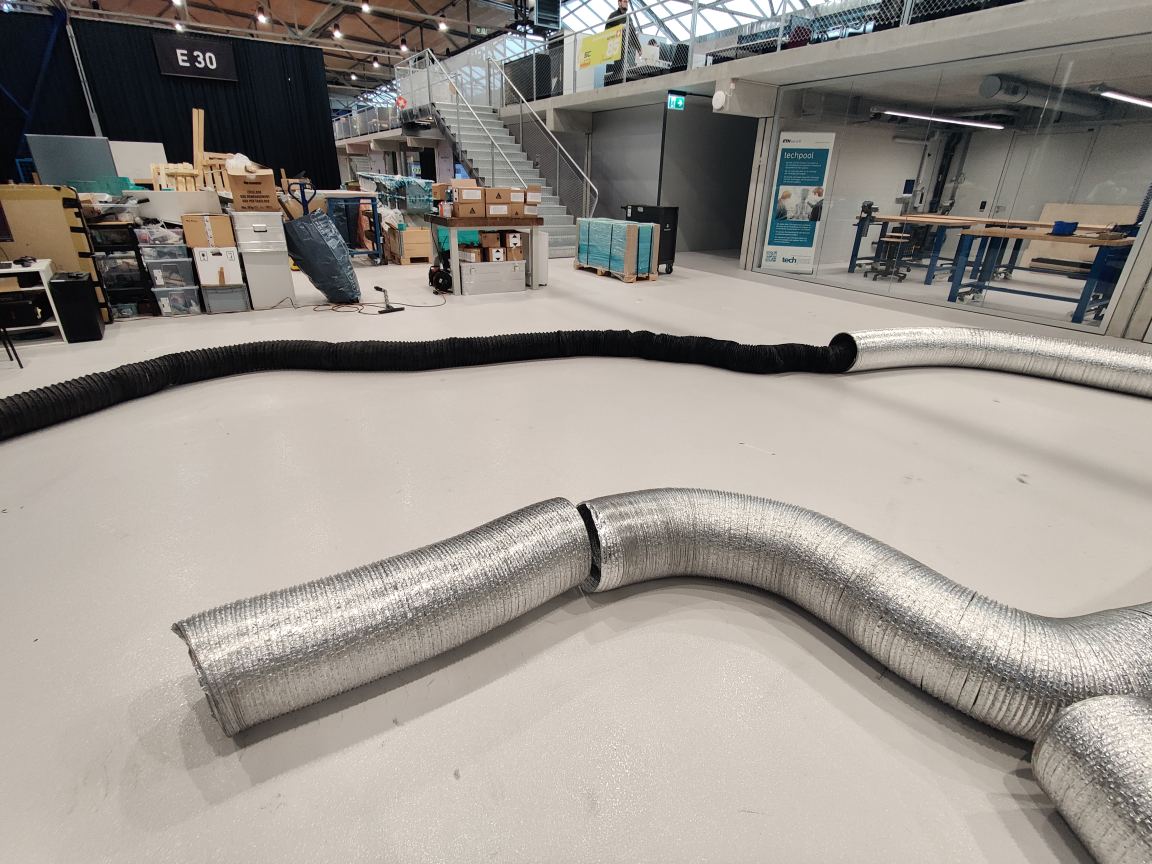}
    \end{subfigure}\\
    \vspace{3pt}
    % \hspace{1pt}
    \begin{subfigure}{0.19\linewidth}
    \includegraphics[width=\linewidth]{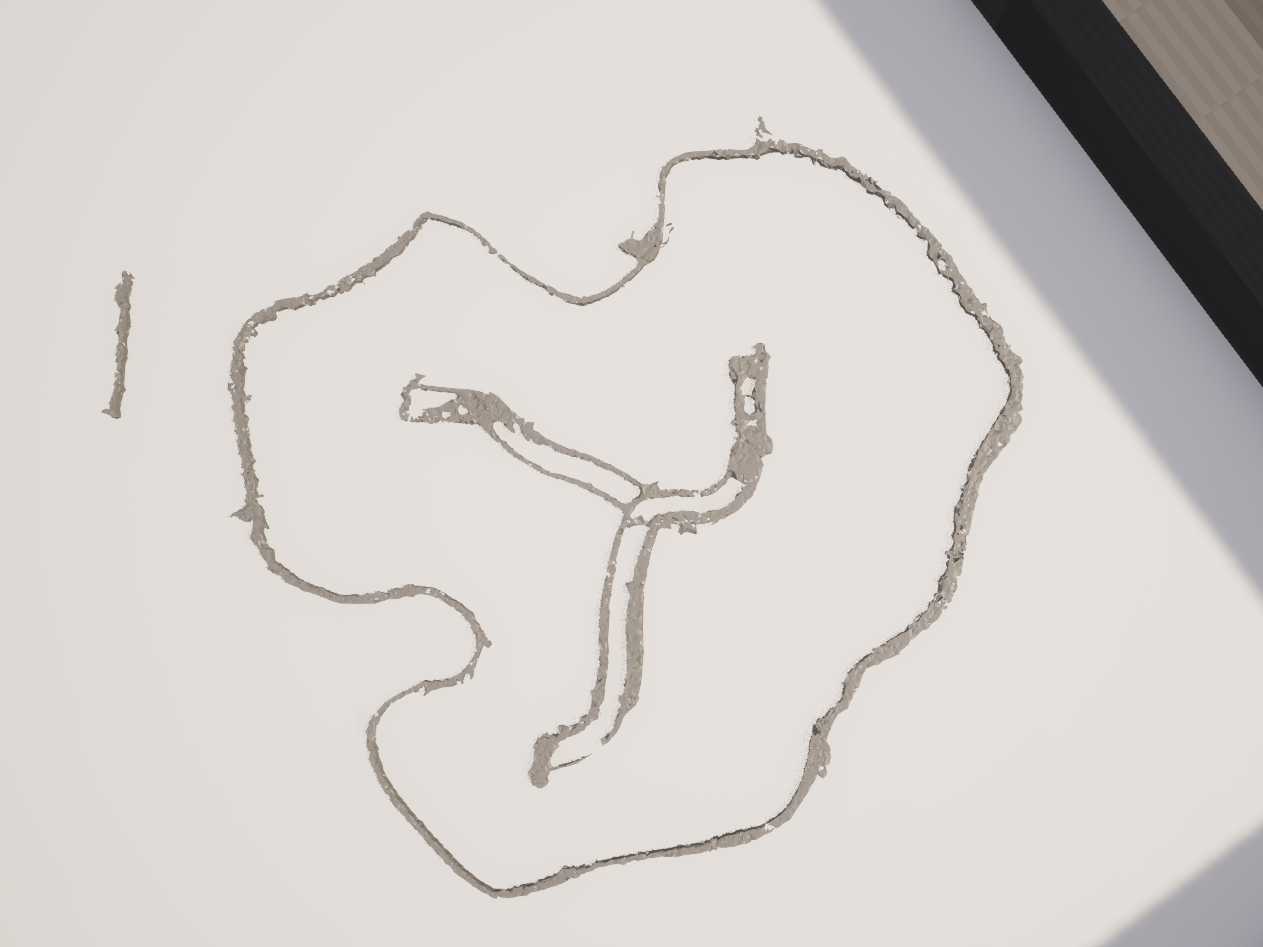}
    \end{subfigure}
    \begin{subfigure}{0.19\linewidth}
    \includegraphics[width=\linewidth]{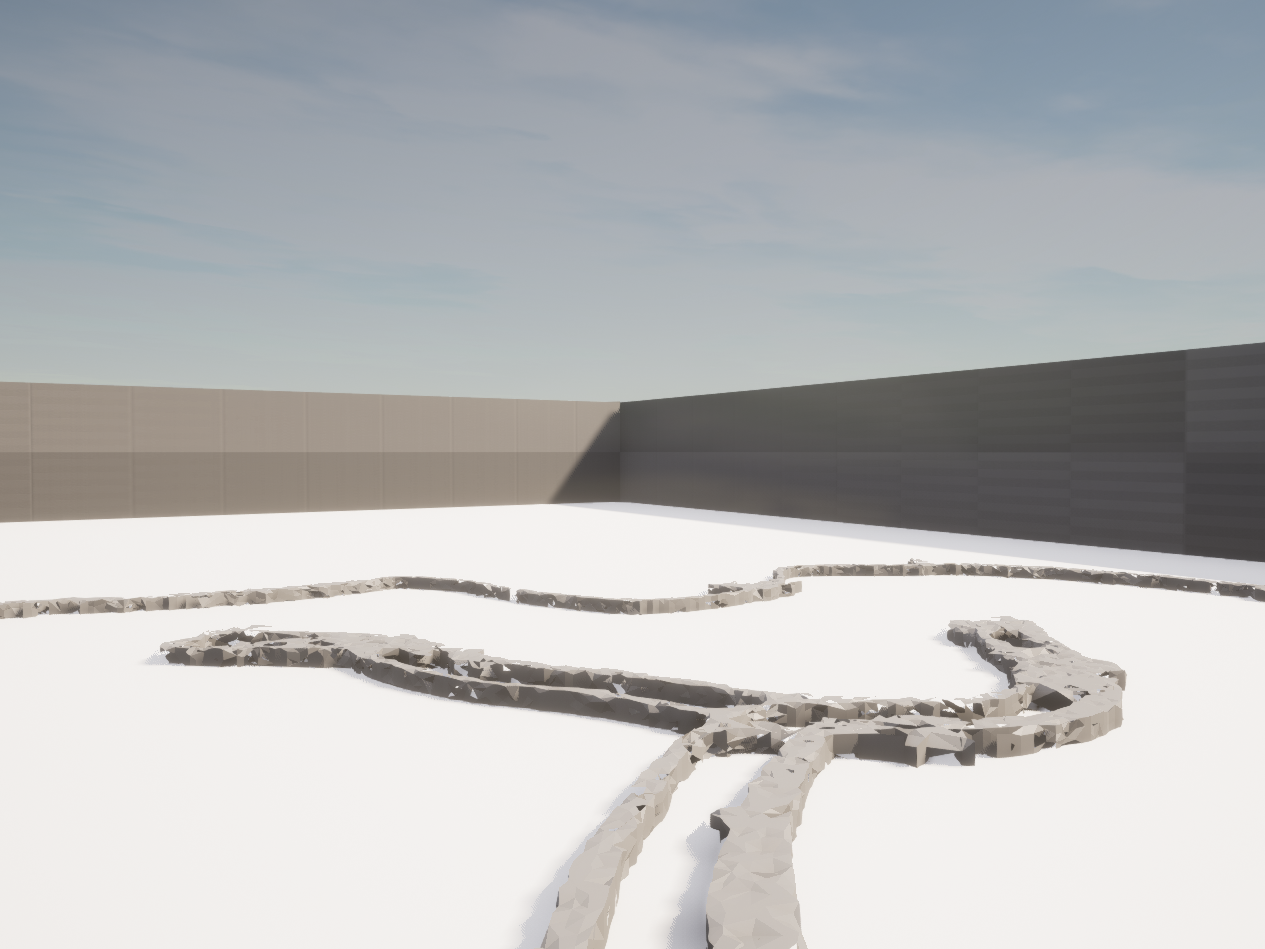}
    \end{subfigure}
    \begin{subfigure}{0.19\linewidth}
    \includegraphics[width=\linewidth]{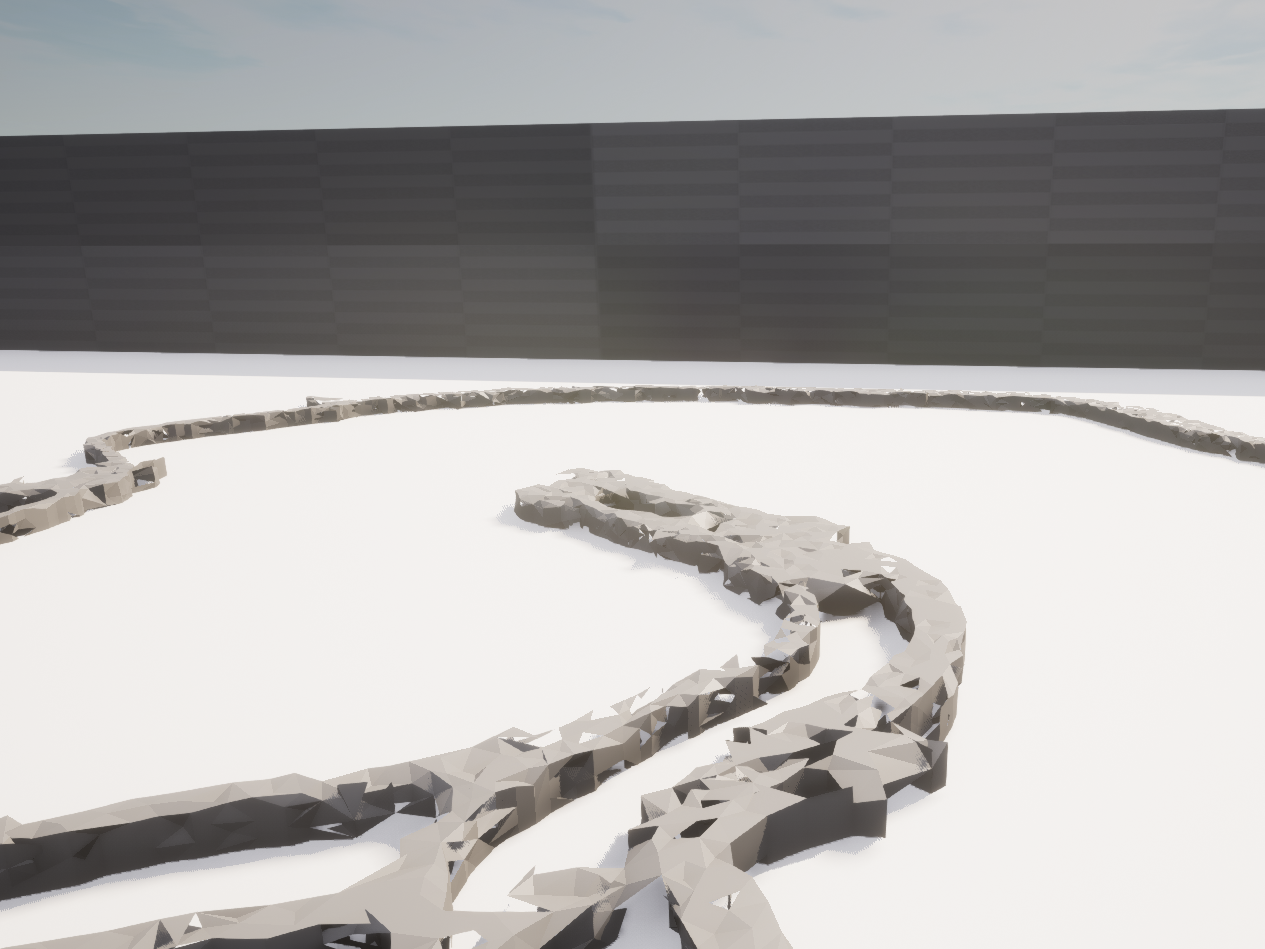}
    \end{subfigure}
    \begin{subfigure}{0.19\linewidth}
    \includegraphics[width=\linewidth]{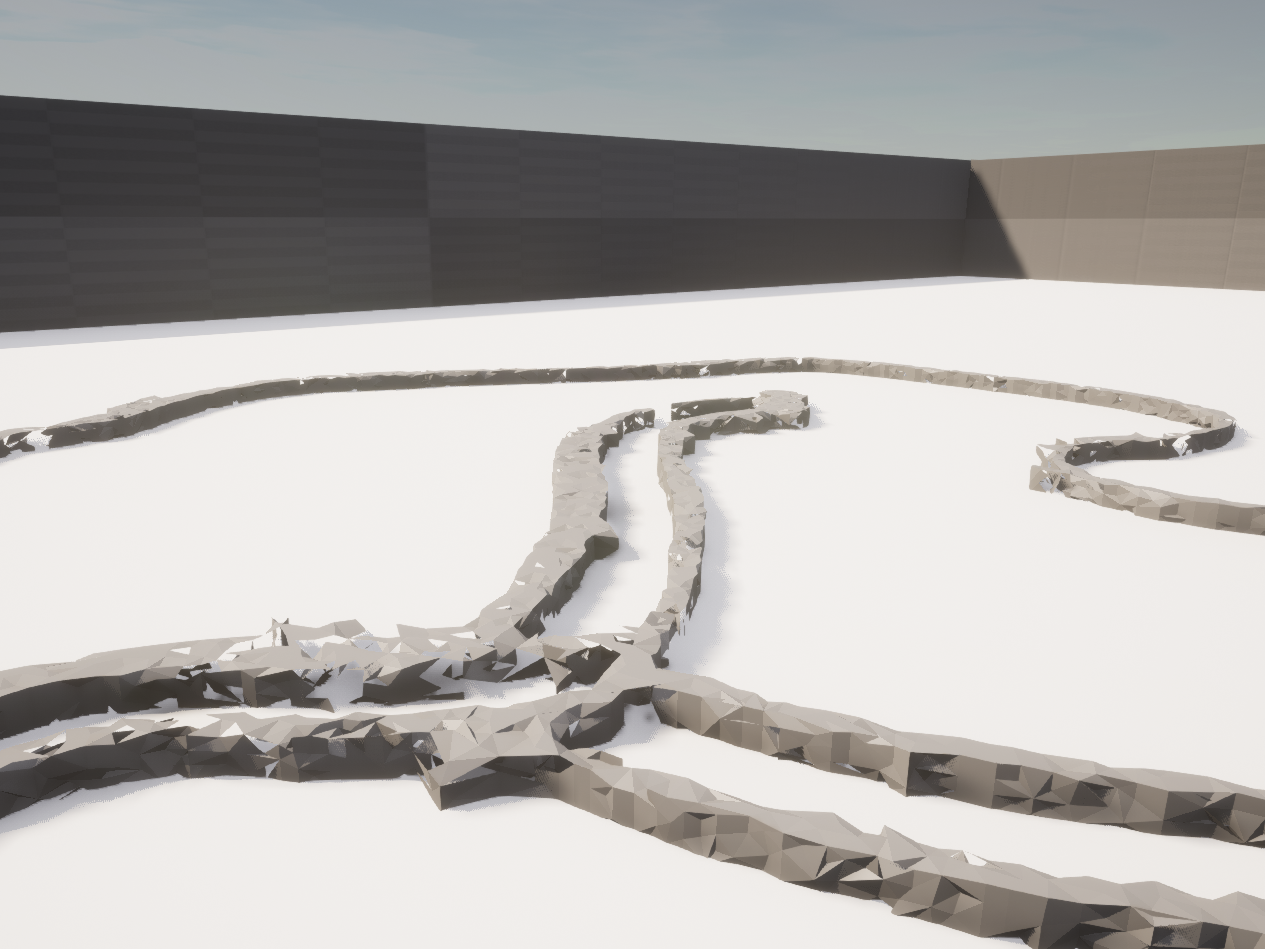}
    \end{subfigure}
    \begin{subfigure}{0.19\linewidth}
    \includegraphics[width=\linewidth]{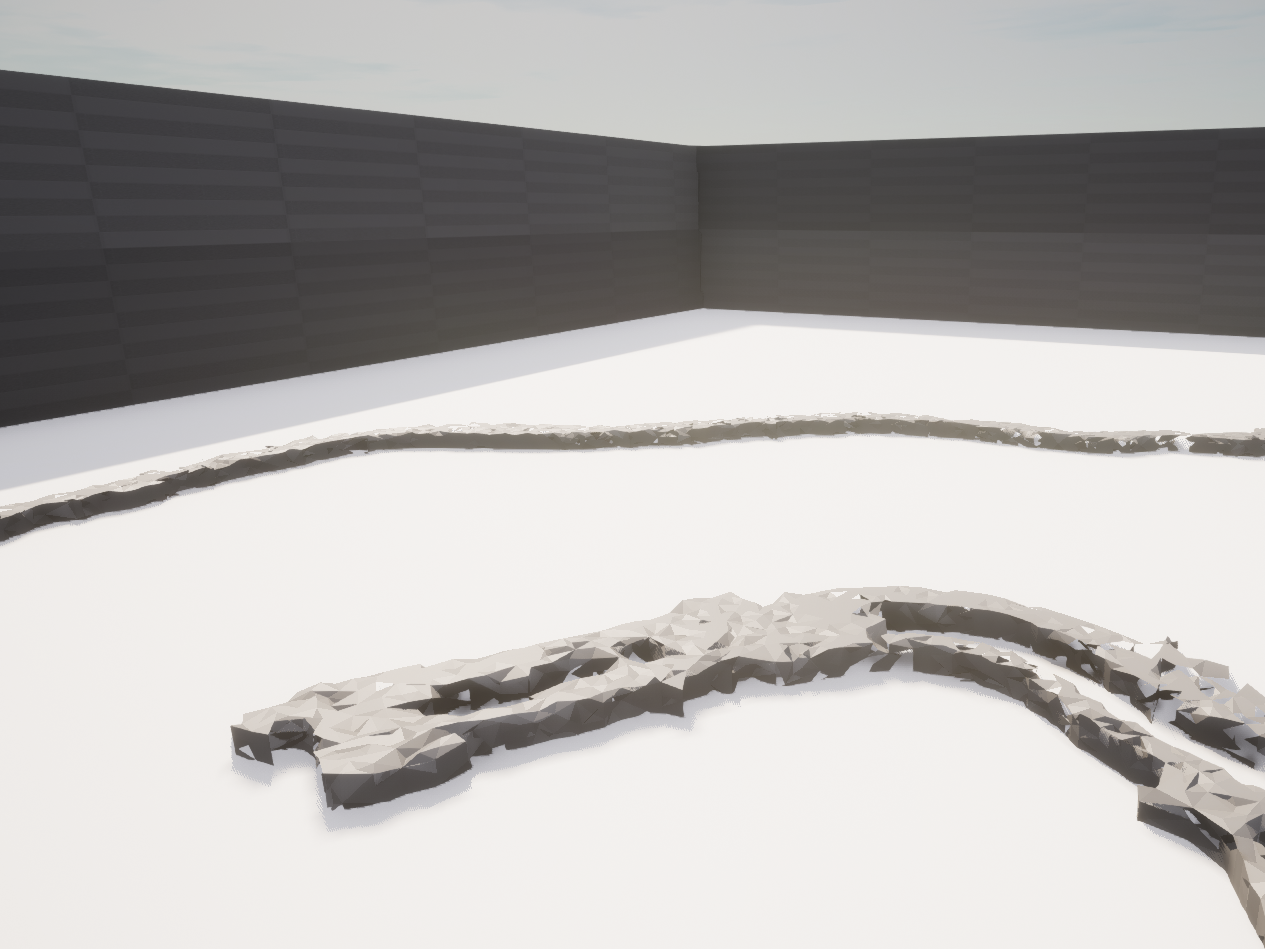}
    \end{subfigure}
    \caption{Top: Different parts of the real environment. Bottom: The resulting digital twin recreated from real-world sensor data.}
    \label{fig:carlamap}
\end{figure*} %twin comparison

\subsection{Modular Dynamics Interface}
\label{sec:dynamics}
Given that the dynamics in CARLA are tailored for urban vehicles, an interface has been introduced in \gls{rcarla} to support the integration of alternative dynamics engines. This interface, implemented using \gls{ros}, processes the updated state $\mathbf{x}_t = [x, y, \psi]$ at each timestep $t$, which includes the vehicle's position $(x,y)$ and orientation $\psi$. The new pose is transmitted to CARLA, enabling it to render the vehicle at the updated location and generate the corresponding sensor data. This configuration allows for the seamless incorporation of any dynamics model capable of computing the current state $\mathbf{x}_t$ based on the sequence of prior states $\mathbf{x}_{0:t-1}$ and the current system input $\mathbf{u}_t$, which consists of the desired longitudinal acceleration $a_t$ and steering angle $\delta_t$. Note that the default dynamics of CARLA do not allow for operation of 1:10 scaled vehicles, hence the proposed dynamics injection are essential if scaled autonomy is desired.

\subsubsection{Dynamic Single-Track Model}
\label{sec:single-track}
To enhance the dynamic simulation, a dynamic single-track model is used \cite{commonroads}. The standard kinematic single-track model computes the updated state of the racecar solely based on geometrical considerations and neglects slip. The dynamic single-track model accounts for this by incorporating the tire forces, $F_{f,r}$ for the front (f) and rear (r) tires, in the state update formula. The state $\bm{x}$ consists of the vehicle's global pose in 2D $x,\,y$, its orientation $\psi$, the longitudinal velocity $v_x$, the lateral velocity $v_y$, and the angular velocity around the z-axis $\dot{\psi}$. The state is updated based on the system inputs at time $t$. The state update is computed as follows:

\begin{equation}
\dot{x} = 
\begin{bmatrix}
    \dot{x} \\
    \dot{y} \\
    \dot{\psi} \\
    \dot{v}_x \\
    \dot{v}_y \\
    \ddot{\psi}
\end{bmatrix}
= 
\begin{bmatrix}
    v_x \cos(\psi) + v_y \sin(\psi)\\
    v_x \sin(\psi) + v_y \cos(\psi)\\
    \dot{\psi}\\
    a_x + \frac{1}{m}(-F_{f} \sin(\delta)) + \dot{\psi} v_y\\
    \frac{1}{m} (F_{f} \cos(\delta) + F_{r}) - \dot{\psi} v_x \\
    \frac{1}{I_z} ( F_{f} l_f \cos(\delta) - F_{r} l_r)
\end{bmatrix}
\end{equation} %pacejka-model

The tire forces $F$ are calculated using the Pacejka tire model as presented in \cite{pacejka1991shear}, this model allows for easy adaptation to different platforms, as it only requires basic vehicle information, such as the mass or height of the center of gravity, along with the corresponding Pacejka tire parameters.

\subsection{Opponent Simulation}
\label{sec:opp}
The opponent simulation module in \gls{rcarla} manages the state updates of simulated opponents. Each opponent requires a predefined reference trajectory composed of waypoints, where each waypoint specifies the pose $[x_i, y_i, \psi_i]$ and the corresponding longitudinal velocity $v_i$. The module advances the opponent along the trajectory by employing linear interpolation between consecutive waypoints $i$-th and the $i+1$-th, using the specified velocity $v_i$ to compute the subsequent pose.

\section{RESULTS}
To assess the effectiveness of the proposed simulator, its \emph{Sim-to-Real} gap is analyzed. This evaluation is conducted for the proposed dynamics model and compared against the CARLA dynamics, as detailed in \cref{sec:dynresults}. Additionally, the \emph{Sim-to-Real} gap is examined for the digital twin creation and sensor simulation, with the results presented in \cref{sec:sensres}.

\subsection{Racecar Dynamics Simulation}
\label{sec:dynresults}
For the analysis of how close to real the proposed dynamics are, two autonomous racecars are used to record laps driven autonomously in different environments with different friction levels. The used vehicles are:

\begin{enumerate}[I]
    \item A 1:10 scaled F1TENTH car with a \emph{Hokuyo UST-10LX} \gls{lidar} and a VESC \gls{imu} and odometry sensor.
    \item A 1:2 scaled gokart with a \emph{Velodyne HDL-32} \gls{lidar}, a \emph{Microstrain 3DM-GX5-AHRS} \gls{imu} wheel, and steer encoders.
\end{enumerate}

For this evaluation, the ForzaETH racestack \cite{baumann2024forzaeth} is used. It uses Cartographer \cite{Hess2016}, a \gls{lidar}-based \gls{slam} system to localize in a pre-recorded map and follows a reference trajectory using a model-based controller. It is deployed on the two cars and used to drive in different environments used for \cref{tab:winti} and \cref{tab:f110}:

\begin{enumerate}[I]
    \item Airfield Track: An F1TENTH racetrack as seen in \cref{fig:carlamap}.
    \item Cellar Track: A second F1TENTH racetrack with a different floor to evaluate the influence of different friction values.
    \item Gokart Track: A gokart racetrack which can be seen in \cref{fig:sys-overview} and \cref{fig:sensors}
\end{enumerate}

Digital twins of the environments are then created from the recorded \gls{lidar} data and the same system is then deployed in the simulator, running on a machine with a \emph{NVIDIA GeForce RTX 3090} GPU and a \emph{Intel  i7-13700K 13th generation} CPU, in three different scenarios denoted in \cref{tab:winti} and \cref{tab:f110} as:

\begin{enumerate}[I]
    \item \textbf{CARLA}: The whole pipeline of the ForzaETH racestack, including state estimation, with the dynamics of CARLA.
    \item \textbf{Ours no SE}: A subset of the racestack, consisting only of the planner and controller modules. The ground-truth output of our proposed dynamics is used as the car state.
    \item \textbf{Ours}: The whole pipeline, including state estimation, with our proposed dynamics
\end{enumerate}

In all scenarios, the lap time, along with the average and maximum lateral deviations from the reference trajectory over a complete lap, were recorded and compared to real-world experiments. In \cref{tab:winti} the differences in these three metrics between the simulations and real-world experiments for both vehicles across all three scenarios are shown. The standard CARLA dynamics are incompatible with the scale of the F1TENTH car due to their reliance on a custom wheel and suspension setup tailored for each vehicle. This setup is unsuitable for cars of such small size, making it impossible to simulate the F1TENTH car within CARLA. The corresponding values are therefore marked as \emph{n/a}. This makes it impossible to give a quantitative analysis of the car dynamics for the F1TENTH platform. However, in the case of the gokart platform, both dynamics systems can simulate the car dynamics. The comparison of our proposed dynamics and the CARLA dynamics shows, that our method achieves much lower differences across all metrics. The deviation in lap time difference on the \emph{Gokart Track} with the gokart vehicle between the simulation and the real data is 5.48 seconds for the original CARLA dynamics and 1.03 seconds for our proposed dynamics, which results in an 81\% reduction of lap time deviation $\Delta T_{lap}$. Similar for maximum $\Delta d_{max}$ and average lateral deviation $\Delta d_{avg}$ the reduction is 34\% and 11\%, respectively. 

These results demonstrate that for the F1TENTH car, the proposed method successfully simulates a platform where the standard CARLA simulator fails. Hence the proposed adaptive vehicle dynamics are essential to enable scaled vehicle operation within CARLA, which results in an overall average reduction of \textbf{42\%}  across all metrics indicating a substantial reduction in the \emph{Sim-to-Real} gap.

\begin{table}
    \centering
    \vspace{4.5mm}
    % \resizebox{\textwidth}{!}
    {%
    \begin{tabular}{l|l|ccc}
    \toprule
    \textbf{Car} & \textbf{Scenario} & \multicolumn{3}{|c}{\textbf{Gokart Track}} \\
    \midrule
    & &\bm{$\Delta T_{lap}\downarrow$} & \bm{$\Delta d_{max}\downarrow$} & \bm{$\Delta d_{avg}\downarrow$}\\
    \midrule

     & CARLA & n/a & n/a & n/a\\
    F1TENTH & Ours no SE & 0.21 & 0.21 & \textbf{0.02}  \\
     & Ours & \textbf{0.064} & \textbf{0.02} & 0.03  \\
    \midrule
     & CARLA & 5.48 & 1.12 & 0.206\\
     Gokart & Ours no SE & \textbf{0.459} & 1.171 & 0.544\\
     & Ours &  1.03 & \textbf{0.73} & \textbf{0.184}\\
     % & Ours vs CARLA & 81\% & 34\% & 106\%\\ --> 73%
    \bottomrule
    \end{tabular}%
    }
    \caption{An overview of the absolute difference of lap time ($\Delta T_{lap}$) in seconds, the average lateral deviation over a lap ($\Delta d_{avg}$), and the maximal lateral error ($\Delta d_{max}$) in meters to the autonomous lap of two autonomous racing platforms in the same environment.}
    \label{tab:winti}
\end{table} %gokart map

\begin{table*}[htb]
    \centering
    \vspace{4.5mm}
    \resizebox{\textwidth}{!}
    {%
    \begin{tabular}{l|l|ccc|ccc}
    \toprule
    \textbf{Car} & \textbf{Scenario} & \multicolumn{3}{|c}{\textbf{Airfield Track}} & \multicolumn{3}{|c}{\textbf{Cellar Track}}\\
    \midrule
    & &\bm{$\Delta T_{lap}[s]\downarrow$} & \bm{$\Delta d_{max}[m]\downarrow$} & \bm{$\Delta d_{avg}[m]\downarrow$}
    &\bm{$\Delta T_{lap}[s]\downarrow$} & \bm{$\Delta d_{max}[m]\downarrow$} & \bm{$\Delta d_{avg}[m]\downarrow$}\\
    \midrule
    & CARLA & n/a & n/a & n/a & n/a & n/a & n/a\\
    F1TENTH & Ours no SE & 0.28 & \textbf{0.06} & 0.07 & 0.172 & \textbf{0.02} & 0.026 \\
     & Ours & \textbf{0.02} & 0.11 & \textbf{0.06} & \textbf{0.021} & 0.04 & \textbf{0.006}  \\
    \midrule
    \bottomrule
    \end{tabular}%
    }
    \caption{This gives an overview of the absolute difference of lap time ($\Delta T_{lap}$), the average lateral deviation over a lap ($\Delta d_{avg}$), and the maximal lateral error ($\Delta d_{max}$) to the real world.}
    \label{tab:f110}
\end{table*} %f110 laps

\subsection{Digital Twin and Sensor Quality Analysis}
\label{sec:sensres}

The gokart is limited to the gokart track due to its larger space requirements. However, for the F1TENTH car, experiments were also conducted on the other two maps, \textit{Airfield Track}, and \textit{Cellar Track}, with the same metrics reported in \cref{tab:f110}. These results illustrate how incorporating the entire autonomy stack pipeline in a holistic setting, affects the \emph{Sim-to-Real} gap. The observed differences in average and maximal lateral error overall 3 maps are 0.14 and 0.19 meters lower in the scenario where the whole pipeline was used. Especially the differences in lap time are significantly lower with a reduction of 69\%, 92\%, and 87\% over all F1TENTH racetracks resulting in an average of \textbf{81\%}. This shows that simulating the entire pipeline holistically is meaningful, as it more accurately reflects the real system. Moreover, it highlights the realism of the simulated sensor data, as any significant divergence from real-life data would have resulted in an increased difference in the used metrics.

\subsection{State Estimation Accuracy}

\begin{table}[h!]
    \centering
    \vspace{4.5mm}
    {%
    \begin{tabular}{l|c|c}
    \toprule
     & $\mathbf{\Delta pos [m]}$ & $\mathbf{\Delta \psi[rad]}$\\
    \midrule

    Synthetic & \textbf{0.1467} &  0.2147\\
    Real & 0.1641 & \textbf{0.1632}\\

    \bottomrule
    \end{tabular}%
    }
    \caption{Overview of the \gls{rmse} of the position and orientation output of the localization of \gls{slam} system in a prebuilt map resulting from synthetic and real data.}
    \label{tab:rmse}
\end{table}

To evaluate the quality of the digital twin environment and the sensor simulation, an experiment was conducted to assess the accuracy of state estimation. A dataset was collected on the \textit{Gokart Track}, consisting of \gls{lidar}, \gls{imu}, and wheel encoder data from the gokart. This dataset was used to create a digital twin of the environment, which was subsequently imported into CARLA. The resulting digital twin map is shown in \cref{fig:sys-overview}.
Within this digital twin, a \gls{lidar}-based \gls{slam} system was deployed to generate a map. The \gls{slam} system enables saving the generated map by exporting its internal state to a file, which can later be reloaded for use in localization mode while replaying the same synthetic data. This configuration represents a best-case scenario for the \gls{slam} system, as the data used for mapping and localization are identical. The accuracy of this setup was evaluated by computing the \gls{rmse} of the position and orientation estimates provided by the \gls{slam} system against the ground truth from CARLA. The results for this scenario are labeled as \emph{synthetic} and presented in \cref{tab:rmse}. Additionally, the synthetic data was used to perform localization in a map created from the original real-world data. The errors resulting from this process are labeled as \emph{real}.

The comparison of \gls{rmse} values between the two cases indicates that the errors are closely aligned. The positional differences, $\Delta pos_{synthetic} = 0.1467$ m and $\Delta pos_{real} = 0.1641$ m, result in a discrepancy of only 1.74 cm, while the heading error difference is less than 0.05 radians. These findings validate the accuracy of the digital twin and sensor simulation, demonstrating that synthetic data can be effectively utilized for localization within a map generated from real sensor data.

\section{CONCLUSIONS}
In this work, a new simulator for autonomous racing is presented. This simulator is capable of holistically simulating different platforms and environments and makes it possible to test full-stack software systems for autonomous racing in a closed loop.
It includes an interface for different car dynamic models, enabling adaptation to multiple simulation frameworks, at vastly different levels of accuracy, to fit different requirements.
The proposed approach also includes a module for simulating opponents so that opponent detection and tracking or overtaking algorithms can be tested. Furthermore, it incorporates a digital-twin creation pipeline which makes it possible to generate a synthetic racetrack from real-world data. The \emph{Sim-to-Real} gap is evaluated on different autonomous racing platforms with different scales and sensor setups which are deployed in various real-life environments. This evaluation shows that the proposed dynamic model reduces the gap by 42\% when compared to the system that it builds upon and the proposed digital twin map creation and sensor simulation reduce it by 81\%.
Further research could delve into extending the proposed system to support new real-world platforms, and include and compare the different dynamic simulation frameworks.


%%%%%%%%%%%%%%%%%%%%%%%%%%%%%%%%%%%%%%%%%%%%%%%%%%%%%%%%%%%%%%%%%%%%%%%%%%%%%%%%

%%%%%%%%%%%%%%%%%%%%%%%%%%%%%%%%%%%%%%%%%%%%%%%%%%%%%%%%%%%%%%%%%%%%%%%%%%%%%%%%

%%%%%%%%%%%%%%%%%%%%%%%%%%%%%%%%%%%%%%%%%%%%%%%%%%%%%%%%%%%%%%%%%%%%%%%%%%%%%%%%

\section*{ACKNOWLEDGMENT}
We extend our gratitude to the entire group of Prof. Dr. Emilio Frazzoli at the Institute for Dynamic Systems and Control for their invaluable support. In particular, we would like to thank Matteo Penlington, Jason Hu, and Marcus Aaltonen for their assistance with the test platform and their guidance during the experiments.

%%%%%%%%%%%%%%%%%%%%%%%%%%%%%%%%%%%%%%%%%%%%%%%%%%%%%%%%%%%%%%%%%%%%%%%%%%%%%%%%

{\small % or \footnotesize, \scriptsize, etc.
\bibliographystyle{IEEEtran}
\bibliography{main}
}
\end{document}